\documentclass{llncs}

\usepackage{amsmath}
\usepackage{amsfonts}
\usepackage{graphicx}
\usepackage{subfig}
\usepackage{tabularx}
\usepackage{booktabs}
\usepackage{qtree}
\usepackage{array}
\newcolumntype{P}[1]{>{\centering\arraybackslash}p{#1}}

\usepackage[usenames,dvipsnames]{color}
\usepackage[linguistics]{forest}

\begin{document}

\mainmatter

\title{LMFD: Latent Monotonic Feature Discovery}
\titlerunning{MDL-based Analysis of Time Series at Multiple Time-Scales}

\author{Guus Toussaint\inst{1}\orcidID{0009-0008-7801-7212} \and Arno Knobbe\inst{1}\orcidID{0000-0002-0335-5099}}
\authorrunning{Guus Toussaint et al.} 
%
%
\institute{LIACS, Leiden University, the Netherlands\\
\email{g.toussaint@liacs.leidenuniv.nl}
}
\maketitle              
\setcounter{footnote}{0}

\begin{abstract}
Many systems in our world age, degrade or otherwise move slowly but steadily in a certain direction.
When monitoring such systems by means of sensors, one often assumes that some form of `age' is latently present in the data, but perhaps the available sensors do not readily provide this useful information.
The task that we study in this paper is to extract potential proxies for this `age' from the available multi-variate time series without having clear data on what `age' actually is.
We argue that when we find a sensor, or more likely some discovered function of the available sensors, that is sufficiently monotonic, that function can act as the proxy we are searching for.
Using a carefully defined grammar and optimising the resulting equations in terms of monotonicity, defined as the absolute Spearman's Rank Correlation between time and the candidate formula, the proposed approach generates a set of candidate features which are then fitted and assessed on monotonicity. 
The proposed system is evaluated against an artificially generated dataset and two real-world datasets.
In all experiments, we show that the system is able to combine sensors with low individual monotonicity into latent features with high monotonicity.
For the real-world dataset of InfraWatch, a structural health monitoring project, we show that two features with individual absolute Spearman's $\rho$ values of $0.13$ and $0.09$ can be combined into a proxy with an absolute Spearman's $\rho$ of $0.95$.
This demonstrates that our proposed method can find interpretable equations which can serve as a proxy for the `age' of the system.
\end{abstract}

\section{Introduction}
Many systems in our world age, degrade, or slowly but steadily move in a certain direction. For example, a highway bridge slowly degrades during its lifetime, a cyclist in the Tour de France will tire over the course of a long stage, and the battery charge of an electric vehicle will deplete as it drives. While in the last example the continuous tracking of the state of charge is fairly doable, in many other applications, the actual `age' of the system may be hidden, and only latently expressed in any data measured about the system. Often, any available sensors will capture the easily measurable information that is often of dynamic nature (what is currently happening in and around the system?), but the actual quantity of interest is much harder to obtain. For example, there can be plenty of measurements for the elite cyclist, including their heart rate, power output, skin conductivity, etc., but the measure of fatigue is hard to define, let alone measure by means of a practical, non-invasive sensor. Still, by looking into the subtle interplay between sensors and how this develops, one might be able to define a data-driven proxy for the quantity of interest. For example, the non-trivial relationship between the current power output of the rider and their heart rate can already provide information about this fatigue (more tired riders will have a higher heart rate for a given power output).

As a motivating example from civil engineering, consider the large concrete highway bridge equipped with a sensor network from the InfraWatch project \cite{Vespier2011,Vespier2012,Vespier2013}, which features 145 sensors measuring at a frequency of 100 Hz. While this collection of sensors is producing substantial amounts of detailed data on a daily basis, the majority of the information is actually not addressing the central quantity of interest, which is the state of health of the bridge. Since concrete bridges generally don't repair themselves over their lifetime of say 50 years, the state of health can be thought of as an accumulation of micro-fissures that collectively constitute the health of the bridge, a quantity that monotonically increases. At the same time, most of the sensors attached to the bridge are primarily sensitive to dynamic changes of the bridge, which are mostly due to elastic changes in the bridge such as increased traffic on the bridge or an elevated temperature due to solar radiation. Across the board, the sensors on the bridge are at least two orders of magnitude more sensitive to dynamic, short-term effects such as heavy trucks, than they are to long-term gradual changes. Still, it is expected that latently hidden in this data lie clues as to the monotonic degradation of the structure.

While individual sensors generally will not show any obvious progressive trend, it might well be that \emph{combinations of sensors} do exhibit this behaviour. Since we do not initially know which sensors to combine and what analytic functions need to be employed to combine them, we have the challenge of \emph{discovering} the right equation that shows a monotonic trend, including what sensors to involve, what functors to apply, and what constants to insert. Our presented solution to the task at hand is to traverse a search space of canonical equations in a top-down fashion, respecting an upper bound on the complexity of the equations (too complex equations are likely not to be of much use, if simpler equations do a similar job). Each candidate equation in our search will have one or more unassigned constants, so these will need to be fitted to the data. In the end, we judge each equation on the basis of its degree of `monotonicity', which we define as the absolute Spearman's Rank Correlation between the time (or any order index of the data points) and the value of the equation.

The specific choice of (absolute) rank correlation as a quality measure allows us some simplifications in the search space, since syntactically different (but related) equations can be proven to have equal rank correlation. For example, the equation
\begin{equation}
   f(s) = a \cdot s + b 
\end{equation}

\noindent
with constants $a \neq 0$ and $b$ will have the same absolute rank correlation as the simpler function
\begin{equation}
 f'(s) = s,    
\end{equation}

\noindent
which is simply the individual sensor $s$. Adding a constant to a sensor clearly has no effect, and the same is true for scaling the sensor with a constant other than 0. A negative value for $a$ will, of course, affect the rank correlation, but not the \emph{absolute} rank correlation. What this trivial example demonstrates, is that there are equivalence classes of syntactically distinct but similar equations, that allow us to compute the monotonicity of a canonical representative of the equivalence class and derive the monotonicity of the other members trivially. Our presented approach involves a grammar of equations that only produces canonical equations and thus avoids costly evaluation of spurious candidates. Note that more complex, non-canonical equations require more effort to fit and subsequently assess, since they will have more constants and functors (as our example demonstrates), so avoiding these will provide great computation benefits.

Assuming our presented approach produces an equation with sufficient monotonicity on a given dataset, one could argue that this equation doesn't necessarily represent \emph{the} age of the studied system, but merely provides a proxy for it, which is fair. But we will argue that any such proxy, as long is it is sufficiently monotonic, will be of interest, at least as long as no definitive alternative definition of age can be given by different means. For example, a discovered equation of good quality that combines a small set of sensors is interesting as an `explainable' insight, meaning that domain experts will benefit from the specific sensors selected, and how they are combined using different functors and constants. On the highway bridge, it turned out that specific locations and orientations of sensors were amenable to monotonic behaviour and others were much less, which is a valuable insight for the design of future sensor setups on other bridges. Furthermore, a monotonic proxy of `age' is valuable since it may point to periods of time where the progress of age/degradation/fatigue was specifically pronounced. These periods can perhaps -- in a further analysis step -- be correlated with other external sources of information, such as traffic or weather data in the case of the highway bridge.

\section{Related work}

In this work, we aim to detect an explainable long-term trend from a multivariate time series.
The Fourier Transform and time series decomposition are previous approaches for detecting and extracting long-term trends from time series.
The Fourier transform allows for the decomposition of a single time series in its individual components as described by Brigham et al.~\cite{brigham1988fast}.
Although the approach can successfully identify long-term cyclic trends by means of extracting the components with a low frequency, it is not suitable for the application of identifying latent ageing in a system.
Ageing does not always move at a constant pace; some events might cause an increase in the speed of ageing, while others might slow it down. 
Thus, when attempting to extract the age with a Fast Fourier Transform, the age component might be scattered around several frequencies.
Furthermore, while the resulting low-frequency components might be monotonic, they are hard to interpret from an `age' perspective since only the frequency is reported instead of an interpretable equation based on the input sensors.
Time series decomposition~\cite{timeseriesdecomposition1997}, which is the process of decomposing a time series into its trend, seasonality, and noise components, suffers from the same limitation.
Even though both approaches can detect long-term trends in time series, a different approach is required to find interpretable, and thus explainable, proxies for age.

Symbolic regression aims to find an equation that combines the independent variables in such a way as to match the target function.
In traditional regression settings, only the parameters are optimised; in symbolic regression, both the parameters and the equation are discovered, as described by Billard et al.~\cite{billard2002symbolic}.
Symbolic regression has proven useful in discovering and verifying equations present in physical systems~\cite{wang2019symbolic,udrescu2020ai}.
Finding the optimal equation for the target function, both in terms of minimal error and simplicity, can be tackled with various methods.
The most common approaches are genetic programming~\cite{schmidt2009distilling,vzegklitz2021benchmarking}, bayesian methods~\cite{guimera2020bayesian}, and context-free grammars~\cite{markic2013cfg}.
However, the specific task defined in this work is an unsupervised one; thus, the previously mentioned methods are not suitable since they require a target function and thus fall into the supervised learning domain.
We address this research gap by proposing a variant of equation discovery that allows for the optimisation with respect to the absolute Spearman's $\rho$, i.e. the monotonicity, thus replacing the supervised learning approach with an unsupervised alternative. Although in this sense, the core of our approach is different from existing equation discovery approaches, a number of commonalities can be identified, for example the use of a context-free grammar and the separation between search (for finding the equation structure) and optimisation (for fitting the parameters of the equation to the data). However, due to the specific nature of our task, a number of novel ideas are involved, for example the definition of a task-specific grammar.

\section{Background}

In this section, we briefly describe the methods used for generating a set of valid equations and optimizing the constants for maximizing the absolute Spearman's $\rho$.

\subsection{Context free grammars}
\label{sec:cfg}

Context-free grammars are formally defined by Parkes et al. \cite{parkes2008concise} as a set of rules which define valid statements in the described language.
Here we leverage context-free grammars to construct equations, as previously described by Todorovski et al. in \cite{co1997declarative}.
Thus, for equation discovery, a context-free grammar can be used as a recipe to construct these valid equations.
Context-free grammars are often described as a tuple $G = (\mathcal{N}, \mathcal{T}, \mathcal{R}, \mathcal{S})$, which contains the following sets of symbols and operations:
\begin{itemize}
    \item $\mathcal{N}$ contains all non-terminal symbols.
    \item $\mathcal{T}$ contains all terminal symbols.
    \item $\mathcal{R}$ contains the rewrite rules in the form $A \xrightarrow{} \alpha$ where $A \in \mathcal{N}$ and $\alpha \in (\mathcal{N} \cup \mathcal{T})^*$.
    \item $\mathcal{S}$ contains the start symbols.
\end{itemize}

As an example of how a context-free grammar can be used to produce valid equations, we will briefly discuss a simple example.
Consider the following grammar $G_s = (\mathcal{N}_s, \mathcal{T}_s, \mathcal{R}_s, \mathcal{S}_s)$, which produces additions and substractions of two sensors $x_1$ and $x_2$.
To achieve this behaviour, the sets of the grammar $G_s$ are defined as follows:
\begin{itemize}
    \item $\mathcal{N} = \{A, B\}$
    \item $\mathcal{T} = \{x_1, x_2, -, +\}$ 
    \item $\mathcal{R} = \{ A \xrightarrow{} B + B\ |\ B - B$, \newline
    \hspace*{9mm} $B \xrightarrow{} x_1\ |\ x_2$\}
    \item $\mathcal{S} = \{A$\}.
\end{itemize}



\subsection{SMAC3}
\label{sec:smac3}

Our equations contain constants which need to be optimised.
We approach this as a black-box optimization problem where our target function is the equation's absolute Spearman's $\rho$.
To achieve this, we used SMAC3 as described by Lindauer et al. \cite{lindauer-jmlr22a}.
SMAC3 provides a framework to optimize the parameters of arbitrary algorithms.
Given the constants as parameters and the absolute Spearman's Rank Correlation as our target, SMAC3 trains a surrogate model to determine the optimal values for the parameters.

\section{Methodology}

\subsection{Monotonicity}

We define the monotonicity of an individual time series by means of the rank correlation between the timestamp or index and the series in question. For the first of these two, it doesn't matter if we use the actual time of each measure or just any other indication of order, since rank correlation measures merely look at the ordering of data. Furthermore, we will use the absolute value of the rank correlation, since we don't care whether a candidate equation is actually progressing in a positive or negative direction. Equations with a negative rank correlation can be simply negated, to achieve the same monotonicity.

There are a number of measures for rank correlation, but in this paper, we will build on the well-known \emph{Spearman's rank correlation coefficient} \cite{Myers2003}, which is defined as follows:
\begin{equation}
    \rho(X,T) = \frac {\operatorname {cov} (\operatorname {R} (X),\operatorname {R} (T))}{\sigma _{\operatorname {R} (X)}\sigma _{\operatorname {R} (T)}}.
\end{equation}

\noindent
Where $X$ is the time series and $T$ is the index or the timestamp of the series in question.
And $\operatorname {cov} (\operatorname {R} (X),\operatorname {R} (T))$ is the covariance between the rank variables, and ${\displaystyle \sigma _{\operatorname {R} (X)}}$ and 
$\sigma _{\operatorname {R} (X)}\sigma _{\operatorname {R} (T)}$ are the standard deviations of the rank variables. Thus, $\rho$ can be thought of as the regular correlation coefficient between the rank variables. As mentioned, we will produce an algorithm that optimises the absolute value of Spearman's rank correlation: $|\rho|$.

\subsection{Grammar definition}
Our search space of candidate equations is specified by a context-free grammar of modest complexity. 
Due to the modest complexity, we will be able to do an exhaustive examination of the equation space (at least, if not an overly large number of time series is involved). 
Our grammar, which we describe in more detail below, assumes that every equation contains at most two series. 
As such, we can let the grammar generate all equation structures, and then instantiate each of these with all the available sensor pairs. 
Assuming the grammar is able to produce $N$ structures, there will be $\mathcal{O}(Nm^2)$ equations to test, with $m$ being the number of time series in the dataset. 

Our grammar will now consist of equations of at most two series $s1$ and $s2$, combined by means of arithmetic operator and further manipulated by potentially useful functions (such as the exponential function) that are monotonic themselves, and thus do not alter the monotonicity of its argument.

To achieve our goal of a sufficiently large search space, we include the following arithmetical operations: addition, multiplication, and division. 
Note that we don't use subtraction, since all terms will include a constant that can be negated to achieve subtraction by addition.
We also include the operations \emph{exp}, \emph{sigmoid}, and \emph{exponentially weighted moving average} (see Section \ref{sec:ewma}).
Then, we define our grammar to include all possible combinations of the previously mentioned operations.
Our grammar $G_m$ is defined as follows:
\begin{itemize}
    \item $\mathcal{N}_m = \{V\ |\ A1\ |\ A2\ |\ B1\ |\ B2 \ |\ Const\ |\ Add\ |\ Mult\ |\ Div\}$,
    \item $\mathcal{T}_m = \{ c1,\ c2,\ c3,\ c4,\ c5,\ s1,\ s2,\ ewma,\ sigmoid,\ exp,\ +,\ \times,\ / \}$
    \item $\mathcal{R}_m = \{V \xrightarrow{} A1\ |\ A2\ Add\ Const\ B1\ |\ A2\ Mult\ B1\ |\ A2\ Div\ B2$, \newline
    \hspace*{10mm} $A1 \xrightarrow{} s1$, \newline
    \hspace*{10mm} $A2 \xrightarrow{} s1\ |\ s2\ |\ ewma(s1, c4)\ |\ sigmoid(s1)\ |\ exp(c2\ Mult\ s1)$, \newline
    \hspace*{10mm} $B1 \xrightarrow{} s2\ |\ ewma(s2,c5)\ |\ sigmoid(s2)\ |\ exp(c3\ Mult\ s2)$, \newline
    \hspace*{10mm} $B2 \xrightarrow{} s2\ |\ ewma(s2, c5)\ |\ sigmoid(s2)$, \newline
    \hspace*{10mm} $Const \xrightarrow{} c1\ Mult$, \newline
    \hspace*{10mm} $Add \xrightarrow{} +$, \newline
    \hspace*{10mm} $Mult \xrightarrow{} \times$, \newline
    \hspace*{10mm} $Div \xrightarrow{} /$, \newline
    \item $\mathcal{S}_m = \{V\}$
\end{itemize}
The set containing the terminal nodes of our grammar ($\mathcal{T}_m$) consists of $5$ constants, $2$ time series, $3$ functions, and arithmetic operations.
The constants, $\{c1, \dots, c5\}$, are treated as parameters that need to be optimised.
Constants $c1,\ c2,\ c3 \in [-1, 1]$  are used for multiplication with time series data.
By optimizing these parameters, we allow the system to alter the weighting and sign of the component.
By doing so the system can find negative relations between time series even though the minus operation is not directly present in our grammar.
Note that the range $[-1,1]$ is sufficient since the role of a $s1$ and $s2$ can be swapped, such that any degree of imbalance between the two can be corrected with a constant multiplication.
The constants $c4$ and $c5$, which range from $1$ to $n-1$, where $n$ denotes the number of data points in each time series, are used as the $span$ parameter for the EWMA as described in Section~\ref{fig:ewma}. 
The symbol $ewma$ denotes the use of the EWMA function. It always takes two arguments, the first represents the input array and the second represents the $span$ used to calculate the EWMA.
The symbol $sigmoid$ denotes the sigmoid function which is defined as follows:
\begin{equation}
    S(x) = \frac{1}{1 + e^{-x}}.
\end{equation}
The sigmoid function, which is strictly monotonic, allows the input to be scaled between $0$ and $1$, in our case this could reduce the impact of outliers on the overall monotonicity of the time series.
The symbol $exp$ denotes the exponential function which is defined as follows:
\begin{equation}
    f(x) = e^x.
\end{equation}
\noindent
The exponential functions, which is also strictly monotonic, is there to (potentially) exaggerate the positive values, while compressing the negative values, such that $exp(x) \geq 0$.

Our grammar's rewrite rules set ($\mathcal{R}_m$) has been designed to generate all potentially interesting combinations of constants, time series, functions, and operations that are relevant in terms of monotonicity. 
More importantly, it avoids generating equations which are syntactically different but can be proven to have the same rank correlation.
The following forms of filtering are included to eliminate spurious equations from the language of equations:
\begin{itemize}
    \item Non-terminals A1 and A2 vs. B1 and B2 make sure that the first series added to an equation has no scaling constant, but the second one does (in order to do relative scaling).
    \item Addition is treated differently from multiplication and division in that only the former can have a scaling constant. Putting a scaling constant in front of a product or quotient has no effect on its monotonicity: $|\rho(axy)| = |\rho(xy)|$ and $|\rho(ax/y)| = |\rho(x/y)|$.
    \item Multiplication is treated differently from division in the sense that division by an $exp$ term is not allowed. This is because a negative constant inside the $exp$ function can produce the inversion caused by division: $exp(-ax) = 1/exp(ax)$.
    \item Although there is a scaling constant within the $exp$ function, there is no second translational constant inside the function. This is because any addition inside $exp$ can be rewritten as multiplication outside the $exp$ and thus can be achieved by the scaling constant: $exp(x+a) = exp(a) \cdot exp(x) = b \cdot exp(x)$.
\end{itemize}
Additionally, the rewrite rule $V \xrightarrow{} A2\ Div\ B2$ can produce equation $s2\ /\ s2$, which is constant and subsequently has a monotonicity of 0.
Therefore, this specific equation is removed from our final set of equations.

Note that our reduced grammar is also designed to minimize the number of required constants, in order to simplify the task of fitting the constants (with the SMAC3 procedure). Fewer constants means that a lower-dimensional fitting problem needs to be solved, increasing the odds of finding a global optimum. The equation structure with the most constants in our language contains three constants.

\begin{figure}
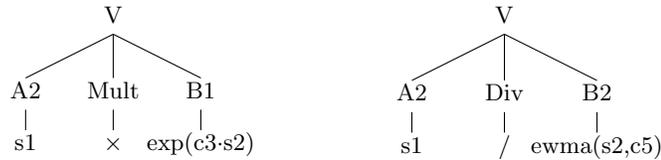

    \centering
\Tree [.V [.A2 s1 ] [.Mult $\times$ ] [.B1 exp(c3$\cdot$s2) ] ]
\hspace*{6mm}
\Tree [.V [.A2 s1 ] [.Div / ] [.B2 ewma(s2,c5) ] ]
    \caption{
        Example of two parse trees derived from grammar $G_m$.
        The produced equations are $s1 * exp(c3 \cdot s2)$ and $s1/ewma(s2,\ c5)$.
        }
    \label{fig:example_parse_trees}
\end{figure}

In total our grammar produces $55$ distinct equations.
Figure~\ref{fig:example_parse_trees} shows the parse trees for two equations which were generated using grammar $G_m$.

\begin{figure}
    \centering
    \Tree [.V [.Mult c6 [[.A2 s1 ] [.Mult $\times$ ] [.B1 exp(c3$\cdot$s2) ] ] ] ]
\end{figure}

\subsection{Convolution operator}
\label{sec:ewma}

Many multi-variate time series describing physical systems will include the notion of \emph{integration over time}, which means its time series' value depends on past values in another time series. Different approaches for this can be found in the literature, including the use of differential equations, recurrent neural networks, and convolution, which is the approach we opt for in this paper. Convolution is a mathematical operator that computes the values of a derived time series as a weighted sum of neighbouring values in the original time series \cite{DSP}.
\begin{equation}
    y(t) = h \ast x(t) =  \sum_{m=- \infty}^{\infty} h(m) x(t-m) 
\end{equation}
As such, convolution can be thought of as a functor in our grammar that takes a time series, and translates it into a derived time series that is potentially more productive than the original time series (just like the \emph{sigmoid} or \emph{exp} functor).

The nature of the convolution operation is determined by the \emph{impulse function} $h(t)$, which determines what weights to give to neighbouring values in the weighted sum. The impulse function is often also referred to as the kernel, which is the terminology we will use in this paper. A kernel that is only based on values from the past, in other words, that has $h(t) = 0$ for all $t < 0$ is called \emph{causal}. As applications of LMFD will be about systems in the physical world, causal kernels are our preferred choice: no information from the future should be involved.

The simplest choice of causal kernel is the \emph{Exponentially Weighted Moving Average} EWMA, which is defined as 
\begin{equation}
    h_{ewma}(m)=e^{-\lambda m}.
\end{equation}
This kernel states that the influence of the last value is the greatest, and gradually declines to zero the further back in history you go. The speed in which this exponential decay occurs is determined by the $\lambda$ parameter. The EWMA kernel is attractive and widely applicable, since it is the solution to a simple first-order differential equation. For example, in \cite{miao2013}, this kernel is used to model the influence of the outside temperature on the temperature of the highway bridge (and thus on the amount of strain) mentioned in the introduction.

\begin{figure*}[t]
   \begin{center}
  \includegraphics[trim=3mm 0 0 0, clip,width=0.9\textwidth]{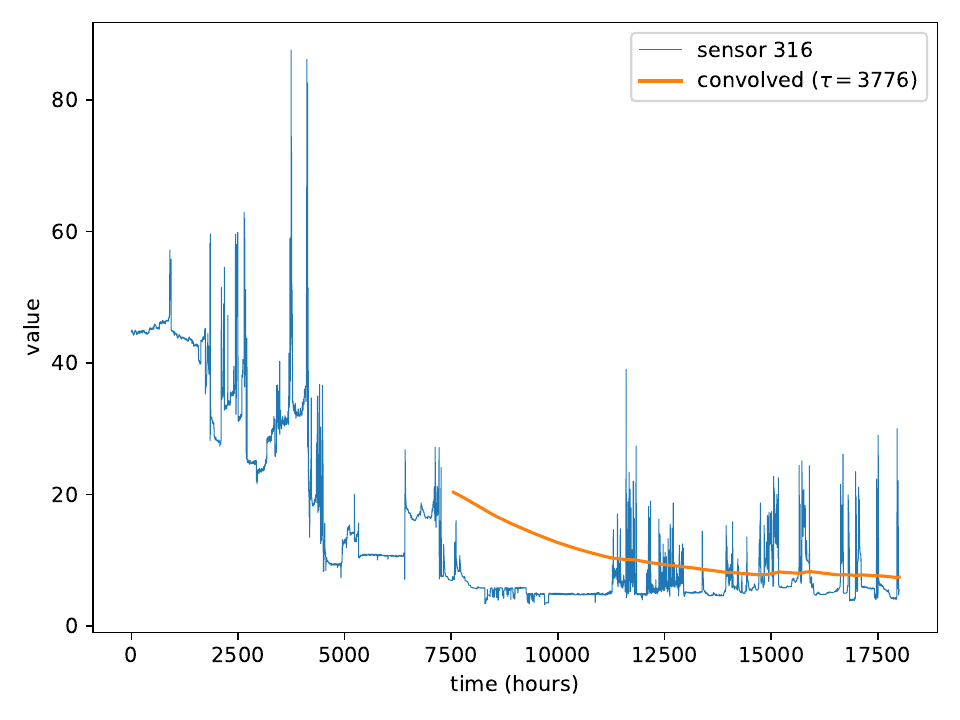}
  \caption{Convolution using EWMA kernel of $\tau = 2402$ hours (about 100 days), applied to two years of data (sensor 316). The derived signal (orange) is rather monotonic, but is not defined for $t < 4805 ( = 2\cdot 2402 + 1)$.}
  \label{fig:ewma}
  \end{center}
\end{figure*}

The Python package we employ for convolution (scipy.stats) conveniently defines the rate of decay in terms of (for example) the center of mass $\tau$, which is related to $\lambda$ as follows: $\tau = 1/\lambda$.
Since in theory, the EWMA kernel has infinite size, any practical implementation will have to assume the kernel is reaching such small numbers that its value can be made zero. For scipy.stats, the kernel is computed over a window of size $span = 2\tau + 1$. Beyond this window, the values of the source time series $x$ no longer play a role. As a corollary of this window, the first $span$ values of the derived time series $y$ are actually undefined (despite the fact that scipy.stats does fill the initial $span-1$ values of the time series with an approximation). For larger values of $\tau$, this means that a considerable part of the convolved time series cannot be computed, as is demonstrated in Figure~\ref{fig:ewma}.

For computing the Spearman's $\rho$ of a time series, having missing values in the time series (as a result of the EWMA operator) is problematic, and computing it over only the available data would introduce a bias. In order to compute $\rho$ in a consistent manner, despite the size of the convolution window, the missing values in the derived time series are imputed with the mean value of the remaining (convolved) data, and $\rho$ is computed over the resulting series. Note that in general, convolution will smooth a time series, and hence make it more monotonic, although only on the latter part of the series. Our imputation approach will counter this unwanted behaviour, and the imputed data will acts as a form of penalty against too large kernels. By imputing a series of constant values we force only part of the time series to be monotonic (the latter part), putting an upper bound on the Spearman's $\rho$, and thus a penalty on the size of the kernel.

\begin{figure*}[t]
   \begin{center}
  \includegraphics[width=0.9\textwidth]{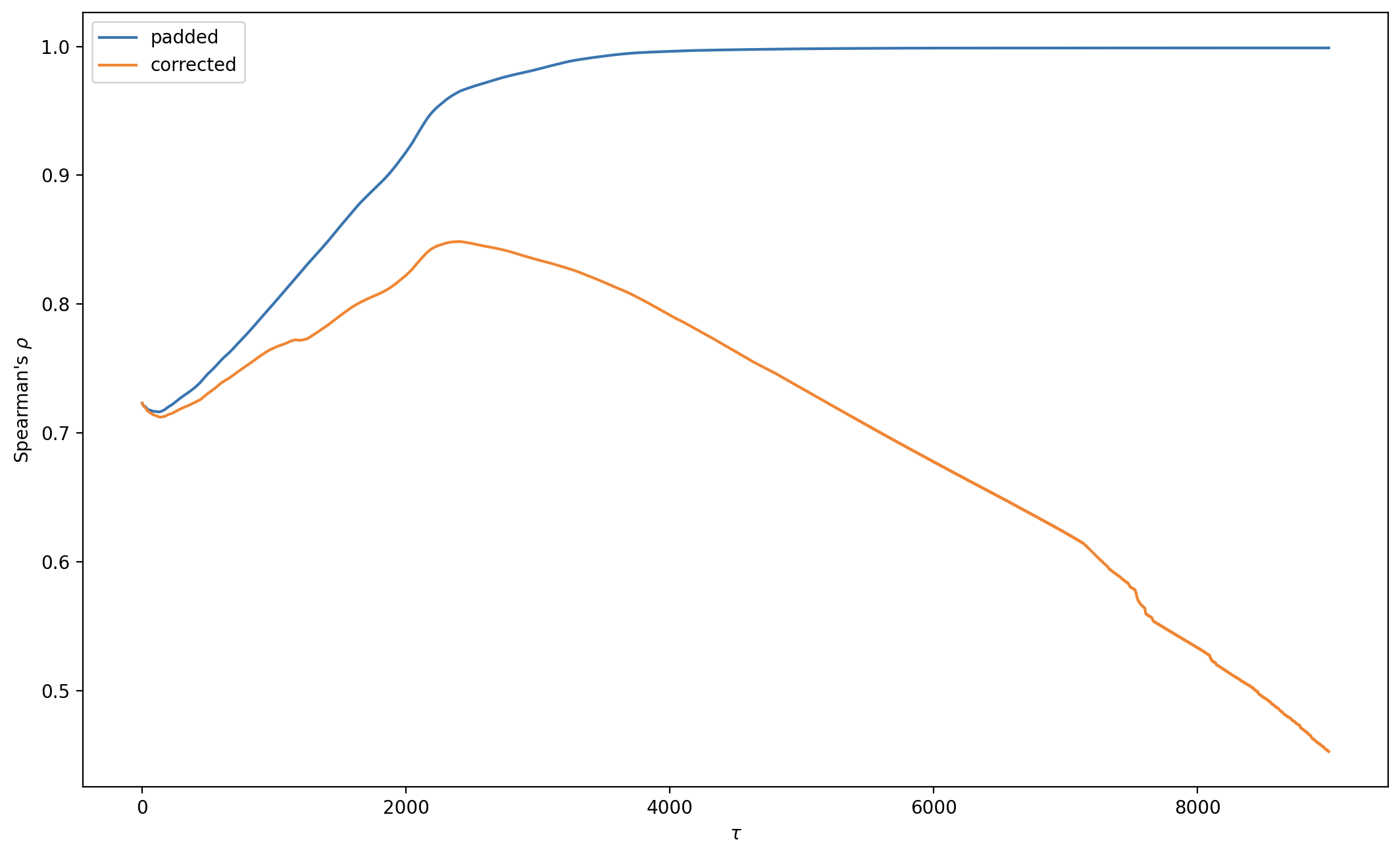}
  \caption{Plot demonstrating the initial rise of $|\rho|$ as the kernel size increases and non-monotonic fluctuations in the data are being smoothed out, to an optimum of $|\rho| = 0.848$, after which the monotonicity declines due to ever smaller parts of the series being available to compute $|\rho|$. \vspace{-4mm}}
  \label{fig:opt316}
  \end{center}
\end{figure*}


\section{Experiments}

\subsection{Setup}
To evaluate our proposed solution for finding latent monotonic features, we use the following three datasets:
First, we have developed our own artificial dataset consisting of two mostly periodic time series that consequently don't show much monotonicity in isolation.
Second, we evaluate the proposed method on two real-world datasets.
The first real-world dataset contains environmental variables measured yearly, and the second contains structural and temperature measurements for a highway bridge in the Netherlands.
Before applying our LMFD algorithm, all time series are first $z$-normalised per series, in order to obtain values centred around 0, with a standard deviation of 1.
All datasets and code for this project are available in the project's repository\footnote{https://github.com/GuusToussaint/LMFD}.

The {\bf artificial} dataset consists of 1000 data points from 2 functions.
The functions are defined as follows:
\begin{equation}
    (1 + sin(\frac{x}{100}))^4 + \mathcal{N}(0, 0.01)
\end{equation}
\begin{equation}
    sin(\frac{x}{100}) + \frac{x}{300} + \mathcal{N}(0, 0.01)
\end{equation}
Where $\mathcal{N}(0, 0.01)$ denotes Gaussian noise with $\mu = 0$ and $\sigma = 0.01$.
These two functions are defined to both be mostly periodic (and hence not very monotonic), but one of the two has a slight positive linear trend superimposed on the larger periodic signal.
Although the shape of the two periodic functions is different, perhaps with the right operations, the two periodic signals cancel each other out, and the trend remains.

Second, the {\bf climate} dataset combines three separate data sources, all recording key environmental variables measured yearly since 1881.
In total, the dataset contains $38$ features, a large proportion of which are rather monotonic by themselves (expressing the unfortunate monotonic trends in CO2 emissions \cite{CO2} and global temperatures \cite{Temp}). 
Since we are interested in discovering a latent monotonic feature, we remove features with an absolute Spearman's $\rho$ greater than $0.8$.
The value of $0.8$ is merely there as a balance between making the task sufficiently difficult (for this paper) while maintaining enough sensors to create a meaningful proxy.

Finally, we evaluate the proposed method on a dataset from the {\bf InfraWatch} project, which contains $17.996$ data points over a period of 2 years, sampled on an hourly basis ($1/3600$ Hz).
For the InfraWatch dataset, we note that a subset of the available sensors, characterised by a certain favourable location and orientation of these sensors, is already fairly monotonic without any additional processing.
Our interest lies in the discovery of interesting combinations of sensors that, perhaps by themselves, are not very monotonic but do show monotonic behaviour when combined in an interpretable equation.
Of the original $119$ sensors, $63$ have a Spearman's $\rho$ $> 0.3$, which we exclude from the analysis.
Using this rather strict threshold, we force the algorithm to focus on the interplay between sensors rather than build on one monotonic feature and add others with minimal weights.
Again, the threshold value is chosen as a balance between making the task sufficiently difficult while maintaining enough sensors to create a meaningful proxy.

\subsection{Results}

\subsubsection{Effects of convolution}
As was outlined in Section \ref{sec:ewma}, any convolution may have a smoothing effect, and if not treated with care, a large kernel may make a series look artificially monotonic. 
In our proposed approach, the missing data at the start, caused by the kernel sticking out `before' the time series, is used as a form of penalty against excessively large kernels.
In Figure~\ref{fig:opt316}, we show the effect of increasing $\tau$ and associated $span$ on $|\rho|$ for the incorrect convolution that pads the missing data and for the corrected convolution that is based on missing data at the start.
Indeed, the padded version finds incorrect proxies, the $|\rho|$ of which approaches $1.0$. On the other hand, our corrected solution finds a reasonably large optimal value of $\tau = 2402$ (this proxy was shown in Figure~\ref{fig:ewma}), but finds lower values of $|\rho|$ for larger $\tau$, due to the increasing amount of missing data at the start for which no monotonicity can be computed.

\subsubsection{Artificial dataset}
\begin{table}[t]
\caption{Top 5 latent monotonic features and the original functions for the artificial dataset.}
\centering
\begin{tabular}{@{} p{0.75\linewidth} P{0.25\linewidth} @{}}
\toprule
Equation & $|\rho|$  \\
\midrule
$ s2 + 0.642*exp(-0.982*s1)$ & $0.9733$ \\ 
$ exp(0.493*s2) + 0.434*exp(-0.841*s1)$ & $0.9696$ \\ 
$ sigmoid(s1) * exp(-0.906*s2)$ & $0.9634$ \\ 
$ exp(0.896*s2) / sigmoid(s1)$ & $0.9634$ \\
$ exp(-0.572*s_{1}) + 0.744*exp(0.874*s_{2}) $ & $0.9587$ \\
... & \\
\midrule
$s2$ & $0.7288$ \\
$s1$ & $0.0753$ \\
\bottomrule
\end{tabular}
\label{tab:top-5-artif}
\end{table}

\begin{figure*}[t]
   \begin{center}
  \includegraphics[trim=3mm 0 0 9mm, clip, width=0.9\textwidth]{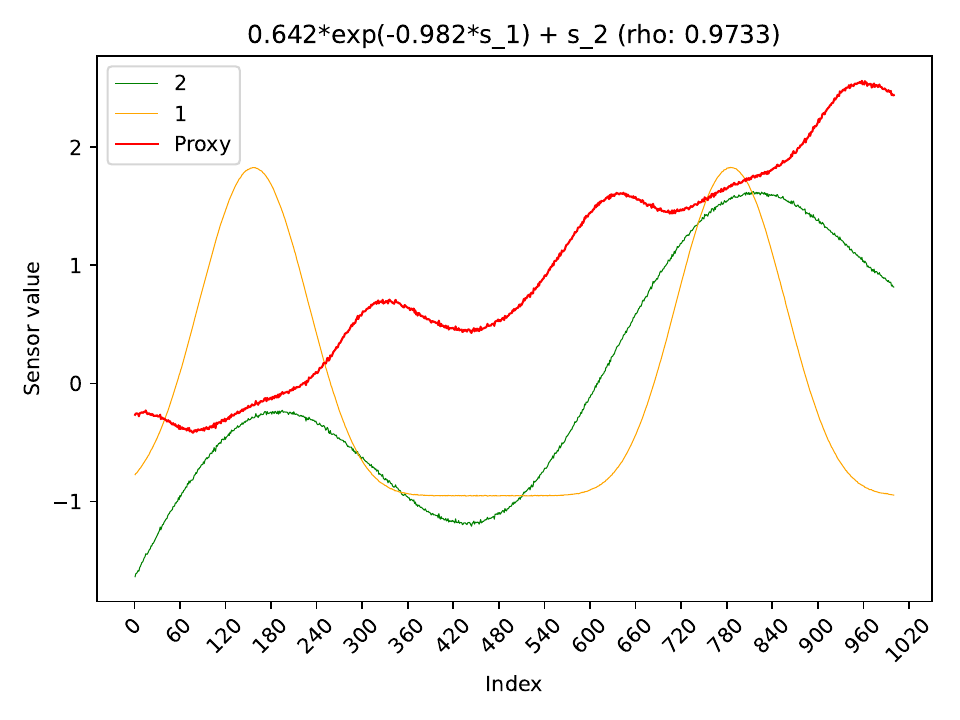}
  \caption{Graph of the latent monotonic feature found for the artificial dataset (red), for $s1$ (orange) and $s2$ (green). The y-axis shows the normalised values reported by the sensors and the proxy. Note that since the proxy combines the sensors, the unit can be undefined if sensors with different units are involved.}
  \label{fig:artif-best}
  \end{center}
\end{figure*}

Table~\ref{tab:top-5-artif} shows the top 5 latent monotonic features in terms of absolute Spearman's $\rho$; for comparison, the raw features are also added.
We observe that all five potential proxies achieve an absolute Spearman's $\rho$ that is greater than $0.95$ while the raw features have an absolute Spearman's $\rho$ of $0.72$ and $0.07$ for $s2$ and $s1$, respectively.
Furthermore, we observe that the best performing latent monotonic feature, in terms of absolute Spearman's $\rho$, achieves a $|\rho|$ of $0.9733$.
Keeping in mind the goal of explainability, this could prove to be an important latent monotonic feature and, consequently, a proxy for the `age' of the system.
Figure~\ref{fig:artif-best} shows a plot of the original features as well as the best-performing monotonic feature.
We notice that the proposed latent monotonic feature successfully leverages the peaks from $s1$ to increase the monotonicity of $s2$, resulting in an overall more monotonic function.
Furthermore, the proxy identifies periods of increased `degradation', which could prove to be useful information if this were to be deployed on a real-world system.

\subsubsection{Climate change dataset}

\begin{table}[t]
\caption{Top 5 latent monotonic features for the climate change dataset extended with the original features used in the latent monotonic features.}
\centering
\begin{tabular}{@{} p{0.75\linewidth} P{0.25\linewidth} @{}}
\toprule
Equation & $|\rho|$  \\
\midrule
$ s_{90S-24S} + -0.595*exp(-0.939*s_{64N-90N})$ & $0.9249$ \\ 
$ exp(-0.584*s_{64N-90N}) + -0.768*s_{90S-24S} $ & $0.9185$ \\ 
$ 0.569*s_{64N-90N} + exp(0.921*s_{90S-24S})$ & $0.9173$ \\ 
$ exp(-0.550*s_{90S-24S}) / sigmoid(s_{64N-90N})$ & $0.9166$ \\ 
$ s_{44S-24S} + -0.604*exp(-0.958*s_{64N-90N})$ & $0.9166$ \\
... & \\
\midrule
$ s_{90S-24S}$ & $0.7962$ \\
$ s_{64N-90N}$ & $0.7800$ \\
$ s_{44S-24S}$ & $0.7769$ \\
... & \\
\bottomrule
\end{tabular}
\label{tab:top-5-climate}
\end{table}

\begin{figure*}[t]
   \begin{center}
  \includegraphics[trim=3mm 0 0 9mm, clip, width=0.9\textwidth]{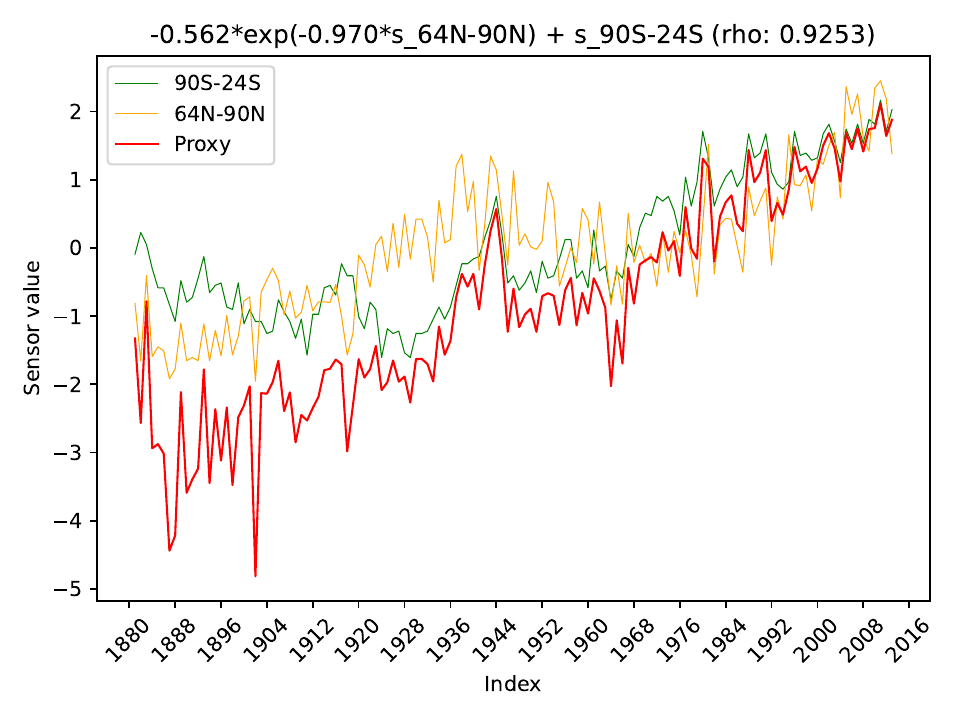}
  \caption{Graph of the latent monotonic feature found for the climate change dataset (red), $s_{90S-24S}$ (green) and $s_{64N-90N}$ (orange). The x-axis shows the date of the recorded value. The y-axis shows the normalised values reported by the sensors and the proxy. Note that since the proxy combines the sensors, the unit can be undefined if sensors with different units are involved.}
  \label{fig:climate-best}
  \end{center}
\end{figure*}

After removing the features that have an absolute Spearman's $\rho$ greater than $0.8$, $13$ of the original $38$ features remain.
For the remaining $13$ features, we execute our LMFD algorithm resulting in $1,612$ equations being evaluated.
Table~\ref{tab:top-5-climate} shows the top 5 latent monotonic features; for comparison, the raw features are added at the bottom.
We observe that from the available $13$ features, only three are used in the top 5 best-performing latent monotonic features.
The other $10$ features apparently do not provide sufficient information to construct a monotonic feature.
We also observe that while the individual features $s_{90S-24S}$, $s_{64N-90N}$ and $s_{44S-24S}$ have an absolute Spearman's $\rho$ of $0.7962$, $0.7800$ and $0.7769$ respectively, all five latent monotonic features have an absolute Spearman's $\rho$ which is greater than $0.9$.
The two features, and the best latent monotonic feature, are shown in Figure~\ref{fig:climate-best}.
Interestingly, $s_{90S-24S}$ measures the yearly global temperature of the southernmost 66 degrees latitude, and $s_{64N-90N}$ the northernmost 26 degrees latitude. Apparently, when regarded individually, they already show a modest rising trend, but combined, the trend is even more pressing.

\subsubsection{InfraWatch dataset}

\begin{table}[t]
\caption{Top 5 latent monotonic features for the InfraWatch dataset extended with the original features used in the latent monotonic features.}
\centering
\begin{tabular}{@{} p{0.75\linewidth} P{0.25\linewidth} @{}}
\toprule
Equation & $|\rho|$  \\
\midrule
$ s_{165} + -0.999 * s_{159}$ & $0.9545$ \\ 
$ exp(0.922*s_{159}) * exp(-0.913*s_{165})$ & $0.9541$ \\ 
$ exp(0.947*s_{165}) * exp(-0.943*s_{159})$ & $0.9539$ \\ 
$ s_{159} + -0.979 * s_{165}$ & $0.9507$ \\ 
$ sigmoid(s_{165}) + -1.000*sigmoid(s_{159})$ & $0.9436$ \\
... & \\
\midrule
$s_{159}$ & $0.1321$ \\
$s_{165}$ & $0.0941$ \\
... & \\
\bottomrule
\end{tabular}
\label{tab:top-5-infrawatch}
\end{table}

\begin{figure*}[t]
   \begin{center}
  \includegraphics[trim=3mm 0 0 9mm, clip, width=0.9\textwidth]{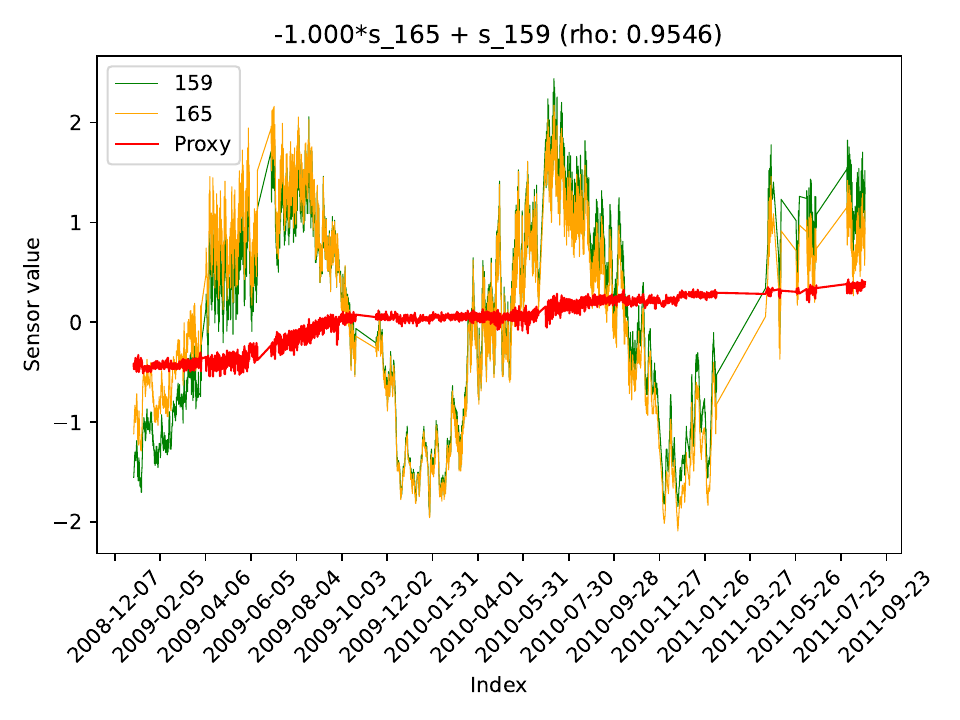}
  \caption{Graph of the latent monotonic feature found for the InfraWatch dataset (red), $s159$ (green) and $s165$ (orange). The monotonicity of the degradation proxy is $|\rho| = 0.9545$. The y-axis shows the normalised values reported by the sensors and the proxy. Note that since the proxy combines the sensors, the unit can be undefined if sensors with different units are involved.}
  \label{fig:infrawatch-best}
  \end{center}
\end{figure*}

The selected $29$ sensors that remain after discarding sensors with an absolute Spearman's $\rho > 0.3$ result in $406$ unique pairs of sensors.
Each unique combination is evaluated by $55$ distinct equations generated by the context-free grammar. 
In total, $44,660$ unique equations are evaluated in order to find the best latent monotonic feature.
Table~\ref{tab:top-5-infrawatch} shows the top 5 best-performing latent monotonic features.
When comparing the absolute Spearman's $\rho$ of the latent monotonic features, we observe a large increase with respect to the raw sensor data.
Furthermore, we observe that a simple weighted subtraction of two raw sensors $s_{159}$ and $s_{162}$ already results in an absolute Spearman's $\rho$ of $0.9545$.
This indicates that the sensor values converge over time which could be a proxy for the `age' of the system.
Sensors $s_{159}$ and $s_{162}$ are both strain gauges situated inside the bridge, as it happens next to each other on the same lane.
This means that for the most part, they will be sensitive to transient changes on the bridge, e.g. caused by traffic passing or a sunny day, in the same way.
As such transient effects should clearly not be part of the age proxy we are looking for, subtracting the two sensors indeed seems like a good idea, because only the differences between how the sensors respond to changes could be relevant.
From our understanding of civil engineering, it indeed makes sense that consider two (or more) sensors jointly would be a good idea, but exactly how to combine these two time series remains unclear.
The equation found by LMFD makes this explicit: simply subtract them with (nearly) equal weights. 
Figure~\ref{fig:infrawatch-best} shows the raw sensor values used in the proxy as well as the proxy itself.
We observe that the sensor values appear to follow the same shape, but as demonstrated by the proxy, their difference tends to decrease over time.
Furthermore, we notice that the decrease in the latent monotonic feature is most profound during the first eight months.

\section{Discussion}

The results on the artificial dataset show that our approach is able to combine two non-monotonic inputs into a latent monotonic feature.
Furthermore, on the real-world datasets, we found a large increase in absolute Spearman's $\rho$ for the discovered latent monotonic feature in comparison to the raw input values.
When analysing the results on the InfraWatch dataset, our approach was able to find the increasing divergence of two similar sensors, which could prove to be an intuitive indication of the degradation of the system as a whole.
We show that even in a challenging environment, where all sensors have an absolute Spearman's $\rho$ less than 0.3, our proposed system is still able to construct a latent feature with an absolute Spearman's $\rho$ of 0.9545. 

We note that our grammar does not include all possible non-monotonic translations, i.e. there are relations between sensors that might bring to light a latent monotonic feature which are not captured in our grammar.
Moreover, the constants added to our grammar are optimised with a black-box approach, resulting in the risk of converging to a local optimum.
Finally, while our choice of only including two separate features in a single proxy does increase the ease of interpretation of the proxy, it limits the degrees of freedom of the system.
In many cases, the degradation of a real-world system is influenced by a multitude of complex features.
Nevertheless, we argue that having an easily understandable measure for this degradation can offer important information and insights to domain experts about the system's overall well-being.

\section{Conclusion}

Many systems suffer from degradation over time, but often, the overall health of a system cannot be directly measured.
We have developed a system that can easily combine two measured features into a latent monotonic feature, which can serve as a proxy for the overall health of the system.
Using an artificial dataset we show that our approach, the LMFD algorithm, works in a theoretical setting, and by using two real-world datasets, we show that the system is able to produce interpretable latent monotonic features from non-monotonic inputs.
For the InfraWatch dataset, we show that two raw input series which have a monotonicity of $|\rho| = 0.13$ and $|\rho| = 0.09$ can be combined into a latent monotonic feature with an absolute Spearman's $\rho$ of $0.95$.

%

\bibliographystyle{splncs04}
\bibliography{refs}

\end{document}